\pdfoutput=1

\documentclass[11pt]{article}

\usepackage[preprint]{acl}

\usepackage{times}
\usepackage{latexsym}
\usepackage{url}

\usepackage[T1]{fontenc}

\usepackage[utf8]{inputenc}

\usepackage{microtype}

\usepackage{inconsolata}

\usepackage{graphicx}

\usepackage{csquotes}
\usepackage{siunitx}
\usepackage{booktabs}

\usepackage{colortbl}
\PassOptionsToPackage{table,xcdraw}{xcolor}
\usepackage{euflag}

\title{Can Out-of-Distribution Evaluations Uncover Reliance on Shortcuts? \\A Case Study in Question Answering}

\author{Michal Štefánik$^\heartsuit$$^\clubsuit$\ \ \ Timothee Mickus$^\heartsuit$\ \ \ Marek Kadlčík$^\clubsuit$\ \ \ Michal Spiegel$^\clubsuit$$^\diamondsuit$\ \ \ Josef Kuchař$^\clubsuit$ \vspace{10pt}\\
        $^\heartsuit$Language Technology, University of Helsinki \vspace{2pt}\\
        $^\clubsuit$TransformersClub @
        Faculty of Informatics, Masaryk University \vspace{2pt}\\
        $^\diamondsuit$Kempelen Institute of Intelligent Technologies}

\begin{document}
\maketitle
\begin{abstract}

A large body of recent work assesses models' generalization capabilities through the lens of performance on out-of-distribution (OOD) datasets. 
Despite their practicality, such evaluations build upon a strong assumption: that OOD evaluations can capture and reflect upon possible \textit{failures} in a real-world deployment.
In this work, we challenge this assumption and confront the results obtained from OOD evaluations with a set of specific failure modes documented in existing question-answering (QA) models, referred to as a reliance on spurious features or prediction shortcuts.

We find that different datasets used for OOD evaluations in QA provide an estimate of models' robustness to shortcuts that have a \textit{vastly} different quality, some largely under-performing even a simple, in-distribution evaluation.
We partially attribute this to the observation that spurious shortcuts are \textit{shared} across ID+OOD datasets, but also find cases where a dataset's quality for training and evaluation is largely disconnected.
Our work underlines limitations of commonly-used OOD-based evaluations of generalization, and provides methodology and recommendations for evaluating generalization within and beyond QA more robustly.


\end{abstract}

\section{Introduction}

Improving the generalization of language models (LMs), i.e., their capability to perform well beyond patterns covered by their limited training data \cite{Chollet2019OnTM,guo2023causal}, presents one of the most important challenges in modern NLP, with direct implications to their practical applicability in a wide variety of tasks.
The most common approach towards evaluating LMs' ability to generalize is to assess their performance on so-called out-of-distribution (OOD) datasets: datasets of the same task(s) but of different origins --- an approach made practical by the large variety of pre-existing datasets for an extensive array of NLP tasks.
We expect a model that generalizes well to achieve high scores on OOD data.
However, as we hold limited knowledge of the properties of datasets, it is difficult to ensure that OOD datasets can comprehensively capture real-world failures to generalize observed in LMs' practical applications.

\begin{figure}[t]
  \!\!\!\!\!\!\!\includegraphics[width=1.08\columnwidth]{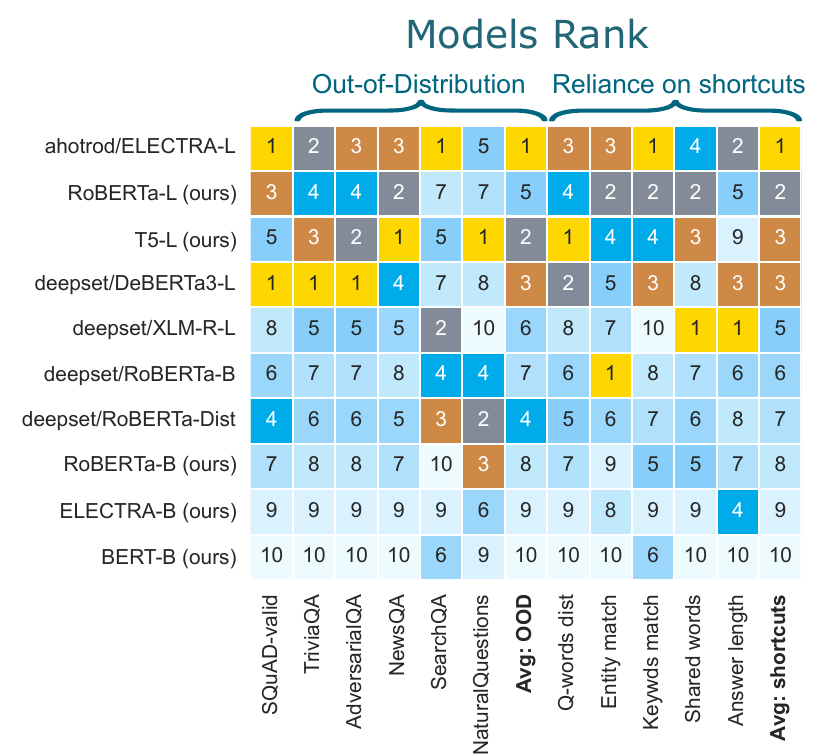}
  \caption{\textbf{How to pick the most robust model?} Ranking of popular QA models by two facets of generalization: Out-of-Distribution evaluations (left) and Reliance on prediction shortcuts/spurious features (right); models are ordered by the average ranking across all spurious features (last column).
  \vspace{-10pt}
  }
  \label{fig:rankings}
\end{figure}

On the other hand, a growing body of work has been documenting and classifying the systematic generalization failures of LMs
as due to
\textit{prediction bias} \cite{utama-etal-2020-mind}, \textit{prediction shortcuts} \cite{mikula2023think}, or a reliance on \textit{spurious features} \cite{pmlr-v139-zhou21g}.
These failures are characterized by a model's over-reliance on a feature that can explain the training data well but is \textit{not} representative of the task overall.

In this case study, we confront these two orthogonal views on generalization --- OOD performance and non-reliance on shortcuts --- by using them to establish two sets of independent rankings for a selection of the most popular QA models.
We focus on extractive QA where previous work provides the most extensive documentation of prediction shortcuts, allowing for a robust assessment.

We find that many OOD datasets can accurately portray models' robustness to shortcuts, but an uninformed selection of OOD datasets can also deliver a ranking that is \textit{not at all} correlated with a reliance on shortcuts, underperforming even traditional in-domain evaluations.
Finally, by assessing reliance on shortcuts for models trained on datasets used as OOD, we show that different QA datasets exhibit the \textit{same types} of shortcuts.
Nevertheless, we find that a dataset's usefulness in uncovering shortcuts does \textit{not} entail that it can be used to train more robust models.


\section{Background}
\label{sec:background}



\paragraph{Evaluating generalization in QA}

Evaluations on out-of-distribution (OOD) datasets are decisively the most popular method for evaluating generalization of language models --- to the point that performance on OOD is often even referred \textit{interchangeably} with the term of generalization \cite{yang2023outofdistribution}.
Within QA,
among many others, \citet{awadalla-etal-2022-exploring} train models on SQuAD \cite{rajpurkar-etal-2016-squad} as in-distribution (ID) and evaluate generalization on TriviaQA \cite{joshi2017triviaqa}, NewsQA \cite{trischler-etal-2017-newsqa}, SearchQA \cite{dunn2017searchqa} or NaturalQuestions \cite{kwiatkowski2019natural} as OOD. \citet{clark2019don} evaluates even shortcut-eliminating method by training on SQuAD and evaluating on AdversarialQA \cite{jia-liang-2017-adversarial} as OOD. \citet{yogatama2019learning} train on SQuAD as ID and evaluate on TriviaQA as OOD.
Despite its known blindspots, SQuAD still remains the default training dataset for a majority of the most popular QA models on HuggingFace.
With a primary objective of improving generalizations, these works \textit{assume} that OOD evaluati\-ons can also uncover reliance on non-representative, spurious features.\footnote{
While we focus on QA, we can easily find such an assumption in other tasks, including NLI \citep{du-etal-2021-towards,korakakis-vlachos-2023-improving} or classification~\cite{yang-etal-2023-distribution}.
}
In this work, we question this assumption and find cases where OOD evaluations are largely \textit{independent} from a reliance on non-representative, spurious features.



\paragraph{Prediction shortcuts}
A complementary yet less prevalent approach to assessing models' generalization aims to exploit functional failures identified in previous models.
One approach towards this goal
%
consists in
identifying models' \textit{prediction shortcuts}, i.e. a reliance on \textit{spurious features} that are not representative for the learned task in general.
Such shortcuts were previously identified in NLI~\cite{nie-etal-2020-adversarial}, in-context learning~\cite{wei2023larger} or question answering~\cite{mikula2023think}.
While these shortcuts are difficult to identify, with their knowledge, we can test the model specifically for the reliance on each of the shortcuts.
This can be done by \textit{constructing} synthetic data \cite{clark2019don} or \textit{subsetting} existing data  \cite{mikula2023think} into subset(s) where we make sure that a specific shortcut is \textit{not} applicable.

In NLI, where the impact of specific prediction shortcuts is widely studied, a common practice is to evaluate on datasets specifically constructed to exploit reliance on shortcuts~\cite{McCoy2019RightFT}.
However, in open-ended tasks including QA or multitask benchmarks such as MMLU~\cite{hendrycks2021measuring}, the applicability of shortcuts across different datasets used for OOD evaluations remains under-studied.

\section{Methodology}
\label{sec:methodology}

\begin{figure}[tb]
  \centering
  \hspace{-15pt}\includegraphics[width=\columnwidth]{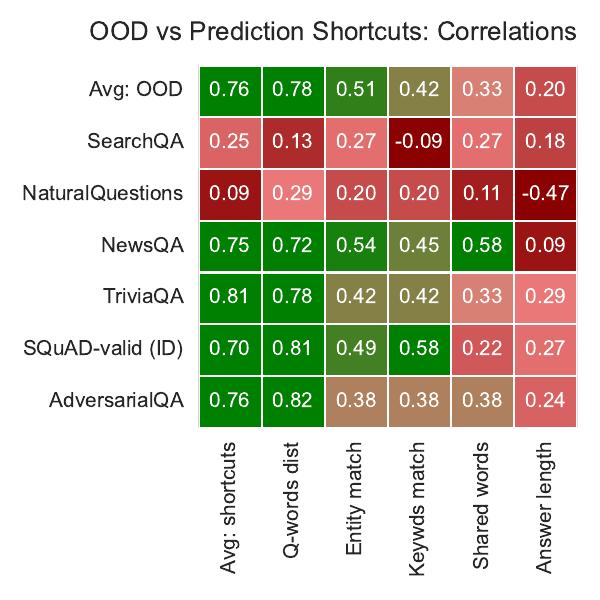}
  \vspace{-10pt}

  \caption{\textbf{Kendall $\tau$ Correlation} (a ratio of changed \textit{orderings}) between the rankings provided by two different facets of generalization: (\textbf{rows}) performance on OOD datasets; and (\textbf{columns}) reliance on prediction shortcuts. First row and column contain an average rankings across all evaluated datasets and shortcuts.
  \vspace{-10pt}}
  \label{fig:correlations_all}
\end{figure}

Our goal is to uncover if, and to what extent can commonly-performed OOD evaluations capture models' failures attributed to models' reliance on prediction shortcuts.
We approach this by \textit{reproducing} several OOD evaluations used in previous work and \textit{comparing} the results of these evaluations with models' measured sensitivity to prediction shortcuts representing previously reported failures modes.



\paragraph{Models}
For our assessments, we aim to pick a set of existing QA models that are the most widely-used in practice.
Towards this goal and with a minor preference for diversity, we pick five among the thirty most-downloaded models on HuggingFace\footnote{\url{https://huggingface.co/models?pipeline_tag=question-answering&sort=downloads}; accessed 21/04/25. The selected models total over 2.4~million monthly downloads.} with the added criterion that they must have been trained on SQuAD only, so as to ensure a fair comparison.
Additionally, to further enhance diversity, we complement this selection by training a set of our own models covering the most popular model families among QA, such that all other training parameters (e.g, early stopping, batch size, learning rate) are fixed across different model families.


\paragraph{ID \& OOD Datasets} Consistent with previous work (§\ref{sec:background}) and the most popular QA models, we use SQuAD \cite{rajpurkar-etal-2016-squad} as our ID dataset unless otherwise noted.
To evaluate robustness, we pick five different OOD datasets, \textit{all} used in prior studies. 
We then use the performances on each of these datasets, as well as their average, to establish a first set of rankings for our models.

\paragraph{Prediction shortcuts}
We create a second set of model rankings by evaluating their reliance on five shortcuts identified in previous work on QA and attested across all models in our study (detailed in §\ref{sec:shortcut}):
shared words \cite{shinoda2021can}, question-words distance \citep{jia-liang-2017-adversarial}, keywords match \cite{clark2019don}, answer length \cite{Bartolo2020BeatTA} and entity match \citep{mikula2023think}.
To quantify the reliance of a given model on each of these shortcuts, we follow the methodology of \citet{mikula2023think}. In practice, we (1)~split the in-distribution validation dataset into two segments based on whether the example (i)~can or (ii)~can \textit{not} be solved by the shortcut; (2)~compute the accuracy on both segments; and (3)~calculate the relative drop in accuracy between the segments.\footnote{
We ensure the statistical significance of our shortcuts through bootstrapped confidence intervals.
}

\paragraph{Metrics}
We assess the agreement between the OOD evaluations and models' reliance on shortcuts by visualizing the ranking obtained by each of these features, and their mutual correlation.
All our reported results employ the exact-match metric.\footnote{
We confirmed on a subset of our evaluations that compared to using F-score, the exact-match metric does not have any effect on the resulting ranking in a majority of evaluations.
See §\ref{sec:appendix_detailed_results} for absolute values for OOD evaluation and shortcut.
}

\section{Results}

Figure~\ref{fig:rankings} displays model rankings as derived from the two approaches for evaluating generalization: (left) OOD evaluations relied upon in previous work (§\ref{sec:background}) and (right) reliance on a set of prediction shortcuts, i.e. a relative drop in models' accuracy when shortcuts are not applicable.
The last column in each group presents an \textit{average} within the group. Models marked as (\textit{ours}) are newly-trained and mutually comparable while other models present the most popular QA models from HuggingFace.

Lower-position rankings are relatively consistent across both OOD and shortcuts.
However, discrepancies become more pervasive at the \textit{top} positions, instructive for picking the \textit{most} robust model.
Here, a majority (3/5) of OOD evaluations do \textit{not} agree on the selection of the most robust model with the average ranking by shortcuts.
Consequentially, picking the `most robust' model based on some OOD datasets (NewsQA and NaturalQuestions) may yield the model with a 23\% larger average dependence on shortcuts.
Noticeably, OOD evaluations on these datasets rank the highest those models that rely on shortcuts \textit{more} than those considered the best by a standard ID evaluation. 


In Figure~\ref{fig:correlations_all}, we compare the different rankings via Kendall $\tau$ pairwise correlations, a metric proportional to the number of pairwise swaps needed to \textit{transform} one ranking into another.
Results reveal that there are two vastly different OOD datasets: NaturalQuestions and SearchQA.
Rankings according to these datasets have \textit{minimal} correlation ($\tau<0.4$) with \textit{each} of the shortcuts, but, as we find, also other with datasets.
Both datasets correlate with the averaged reliance on shortcuts substantially \textit{worse} than the traditional ID evaluations.
On the other hand, some OOD evaluations correlate with average shortcuts' ranking much better than ID or even averaged OOD performance --- which suggests that ranking among a large set of OOD evaluations, assumed as more robust in some of previous works~\cite{awadalla-etal-2022-exploring} need \textit{not} provide a better assessment of robustness than an informed selection of a single OOD dataset.




\subsection{Analyses}

The large discrepancy among different evaluation datasets in their ability to uncover shortcuts raises the question of whether different datasets can indeed exhibit the \textit{same} prediction shortcuts.
If so, models' reliance on shortcuts would not only remain hidden from OOD evaluations but could even \textit{improve} their OOD results.
To answer this, we train new QA models on each of our OOD datasets and assess their reliance on each of our shortcuts. 
Except for the training data, our methodology remains identical to the training of ID models (§\ref{sec:methodology}).
We limit our analyses to the model least reliant on shortcuts, viz. RoBERTa-Large. 

\begin{table}[]
    \centering
    \resizebox{\linewidth}{!}{
    \begin{tabular}{l@{{\,}}l *{6}{S[table-format=2.2, round-mode=places, round-precision=2]}}
        & & {{\rotatebox{90}{\textbf{TriviaQA}}}} & {{\rotatebox{90}{\textbf{Adv'QA}}}} & {{\rotatebox{90}{\textbf{NewsQA}}}} & {{\rotatebox{90}{\textbf{SQuAD}}}} & {{\rotatebox{90}{\textbf{SearchQA}}}}  & {{\rotatebox{90}{\textbf{NQ's}}}} \\
        \toprule
        \textbf{Eval} & ($\uparrow$) & \cellcolor{green!50} 0.87 & \cellcolor{green!40} 0.78 & \cellcolor{green!30} 0.75 & \cellcolor{green!10} 0.70 & \cellcolor{red!30} 0.25 & \cellcolor{red!50} 0.09 \\
        \textbf{Train} & ($\downarrow$) & \cellcolor{red!40} 15.28 & \cellcolor{green!40} 2.79 & \cellcolor{green!30} 3.09 & \cellcolor{green!10} 3.17 & \cellcolor{red!50} 16.21 & \cellcolor{red!5} 3.22 \\
        \bottomrule
    \end{tabular}
    }
    \caption{\textbf{Quality of datasets} for (top) \textbf{evaluation}, i.e. dataset's ability to uncover shortcuts in evaluation (reported in Fig.~\ref{fig:correlations_all}; $\uparrow$ higher is better), and (bottom) for \textbf{training}, i.e. a reliance on shortcuts for models trained on the given dataset ($\downarrow$ smaller is better).
  \vspace{-12pt}}
    \label{tab:eval_vs_training}
\end{table}

In Table~\ref{tab:eval_vs_training}, we report two metrics for each of our datasets: (top row)~the correlation with reliance on shortcuts when \textit{evaluating} with the dataset (identical to Fig.~\ref{fig:correlations_all}), and (bottom row)~the average reliance on shortcuts (a relative drop in accuracy) when \textit{training} on this dataset.
The datasets (x-axis) are ordered based on their correlation with ranking based on the reliance on shortcuts (Fig.~\ref{fig:correlations_all}).

\paragraph{Can different OOD datasets exhibit the \textit{same} prediction shortcuts?}
Table~\ref{tab:eval_vs_training} shows that except for AdversarialQA and NewsQA, models trained on OOD datasets rely on shortcuts of SQuAD similarly or even \textit{more} than a SQuAD-trained model.
In the case of TriviaQA and SearchQA, the drop in accuracy caused by the unavailability of shortcuts is around \textit{five times larger} than that of the SQuAD model.
However, detailed results (Appx.~\ref{sec:appendix_detailed_results}) reveal that even the less-reliant AdversarialQA and NewsQA-trained models exhibit a significant reliance on shortcuts in the case of three and four out of seven inspected shortcuts.
Together, these evaluations provide evidence that datasets used for OOD evaluations in previous work exhibit the \textit{same} types of prediction shortcuts as the training data.

\paragraph{Are bad \textit{training} datasets also bad \textit{evaluation} datasets?}
We showed that different datasets can provide vastly different quality in both training robust models and uncovering reliance on shortcuts.
Consequentially, we may assume that there is a \textit{proportional} relationship between the dataset's ability to train a robust model and to uncover the reliance on shortcuts.
However, relating these two facets in Table~\ref{tab:eval_vs_training} (\textit{Train} vs \textit{Eval} row), we can see \textit{no} clear relation between the dataset's quality for training and evaluation;
For instance, while TriviaQA is the best proxy for evaluating models' reliance on shortcuts, using it as a training dataset delivers a model almost five times more reliant on shortcuts than less reliable SQuAD or NaturalQuestions (NQ's).


These results point to the presence of \textit{other} covariates that determine datasets' quality independently for training and evaluation.
We investigate several potential features, including dataset size, context size, and sample format.
While we do not identify a robust discriminant in training, in evaluation, we find that the least robust evaluations are delivered by datasets with more specific formats containing delimiters of different context sections (SearchQA) or in-context references (NQ's).
We hypothesize that while the context format may not necessarily harm the robustness in training, it may strongly bias the evaluation towards dominantly assessing robustness to the dataset-specific artifacts.

\section{Conclusions}
\label{sec:conclusions}

This paper investigates a discrepancy between OOD evaluations used in previous work in QA as a proxy for generalization, and previously documented failures to generalize characterized by prediction shortcuts.
By ranking a set of popular QA models according to these two facets, we find that datasets previously used for OOD evaluations vastly differ in their capacity to uncover shortcuts.
%

We find that prediction shortcuts are to a large extent \textit{shared} across datasets used for both training and evaluations of generalization.
However, the quality of datasets in uncovering shortcuts is \textit{not} proportional to their capacity to train more robust models, possibly due to dataset-specific features with a different impact on training and evaluation.

We hope our results will inspire future work towards a more systematic selection of OOD datasets, while providing concrete recommendations for OOD datasets selection in QA. Crucially, our findings may also motivate future work in generalization beyond QA to restrain from over-generalized conclusions based on limited OOD benchmarks.

\section*{Acknowledgements}
This work is supported by the Research Council of Finland through project No.~353164 ``Green NLP -- controlling the carbon footprint in sustainable language technology''.
\euflag \ This project has received funding from the European Union’s Horizon Europe research and innovation programme under Grant agreement No. 101070350 and from UK Research and Innovation
(UKRI) under the UK government’s Horizon Europe funding guarantee (grant number 10052546). The contents of this publication are the sole responsibility of its authors and do not necessarily reflect the opinion of the European Union.

\section*{Limitations}

We identify a primarily limitation of our work in a limited scope of prediction shortcuts that we survey. This limitation is conditioned by the scope of (seven) types of shortcuts identified in previous work, among which we identify only five to be significant for most of our evaluated models.
This also restrains us from expanding our case study into other candidate tasks; in NLI, we identify only three known prediction shortcuts~\cite{McCoy2019RightFT}, while in in-context learning, we identified in previous work only a single prediction shortcut to assess (\textit{reliance on label's semantics} uncovered in \citet{wei2023larger}).

We further acknowledge that the database of known prediction shortcuts presents only a small subset of functional failures, where the failures of other categories would certainly be also desirable to capture in generalization evaluations.
This limitation invites future work to assess models for other notorious functional deficiencies as our knowledge of models' functioning will grow, in a methodology similar to ours.

\bibliography{stefanik}

\appendix

\begin{figure*}[t]
  \centering
  \hspace{-40pt}\includegraphics[width=1.0\textwidth]{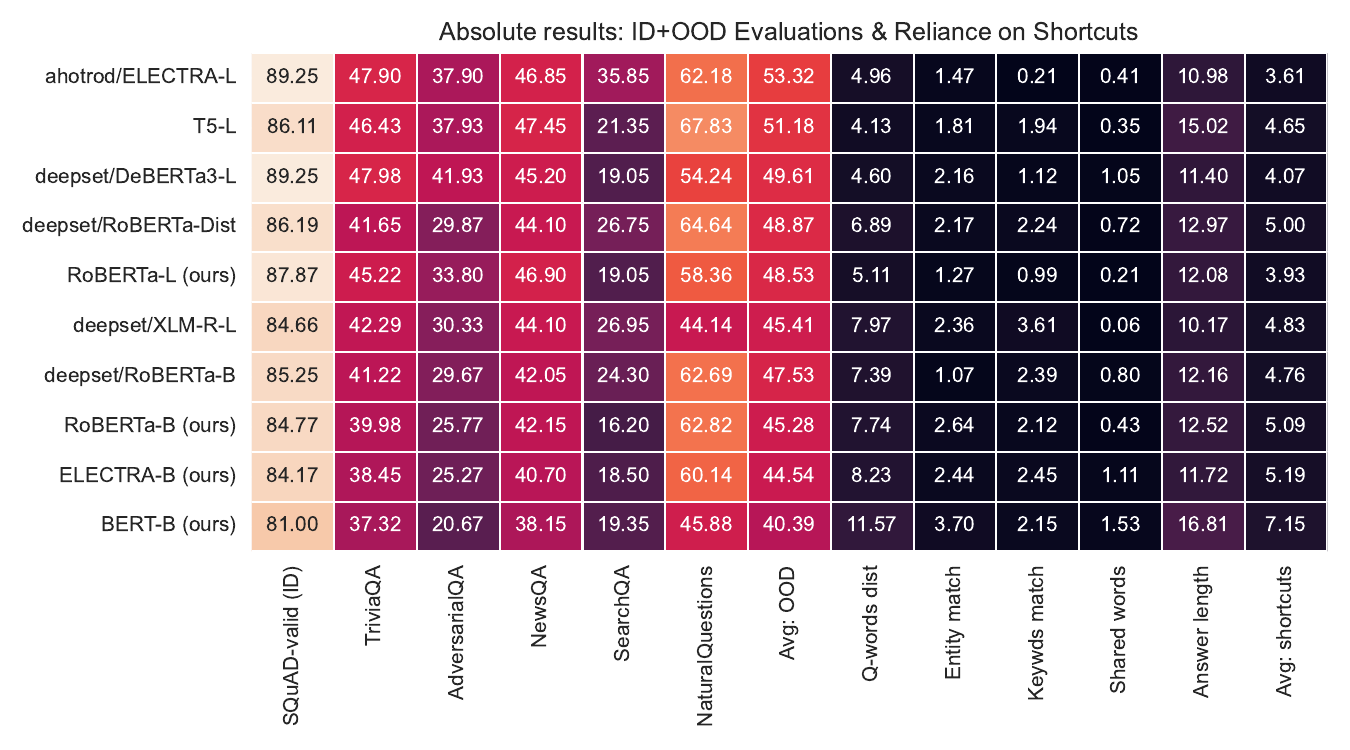}
  \caption{Exact-match results of (i)~OOD evaluations and (ii)~reliance on prediction shortcuts denoting a minimal relative drop in accuracy (in percentage) when facing data where the prediction shortcut is not applicable. See the \textit{Prediction bias} algorithm in \citet{mikula2023think} for further details. These evaluations were used to compute the ranking and correlations in Figs.~\ref{fig:rankings} and \ref{fig:correlations_all}.}
  \label{fig:exact_results_all}
\end{figure*}

\section{Definitions of shortcuts}
\label{sec:shortcut}
We include here a brief overview of the shortcut we study and refer the reader to the original works for relevant details.

\begin{itemize}
    \item \vspace{-7pt}\textbf{Shared words} \cite{shinoda2021can}: Models' reliance on the assumption that correct answer must contain some words from the question.
    \item \vspace{-7pt}\textbf{Question-words distance} \cite{jia-liang-2017-adversarial}: Models' reliance on close proximity of the answer to some of the words contained in the question.
    \item \vspace{-7pt}\textbf{Keywords match} \cite{clark2019don}: Models' reliance on the matching keywords, i.e. low-frequency words between the question and answer.
    \item \vspace{-7pt}\textbf{Answer length} \cite{Bartolo2020BeatTA}: Models' false reliance on a specific word-level length of the answer.
    \item \vspace{-7pt}\textbf{Entity match} \cite{mikula2023think}: Models' reliance on that the answer must contain a first entity matching the type of the question, such as ``Who'', ``Where'', etc.
    \vspace{-5pt}
\end{itemize}

\section{Experimental Details}
\label{sec:appendix}

Our experiments train a set of QA models, separately on each of our training and evaluation dataset.
Towards the goal training models representative for real-world deployment, we perform hyperparameter search within each model family for optimal values of learning rate (including values 1e-6, 2e-6, 1e-5, 2e-5, 4e-5, 5e-5, 2e-4) and batch size (including values 8, 16, 32, 64).
We used early stopping based on evaluation loss based on (in-distribution) validation set of SQuAD, patience=5, evaluations every 2000 updates and a maximum of 5 epochs.
In this configuration, we were able to train each of our 11 trained models under 24 hours on a single Nvidia A40 GPU.
Our training scripts are using HuggingFace Transformers library \cite{Wolf2019HuggingFacesTS}.
Training script can also be found among supplementary materials of this submission.

All our evaluations, including ID and OOD evaluations, employ exact-match metric and on the case of our own-trained models, we verify that the choice of our primary evaluation metric has no effect on the ranking of models in a majority of OOD evaluations, while causing at most two swaps in ranking in other cases.

Evaluations of Reliance on shortcuts follow the methodology of \citet{mikula2023think}; here, we employ the \texttt{isbiased} library from the Authors' referenced GitHub repository\footnote{\url{https://github.com/MIR-MU/isbiased}}.
Following the original methodology, we asssess all reliances on shortcuts consistently using the SQuAD 1.1's standard validation set of the full size.
We include also our evaluation script among the supplementary materials of this submission.

\subsection{Detailed Results}
\label{sec:appendix_detailed_results}

Figure~\ref{fig:exact_results_all} shows the detailed results with the absolute values of out-of-distribution evaluations as well as the relative dependencies on a reliance on surveyed spurious features for all our evaluated models.
The OOD values are listed in exact-match metric, while the dependencies on spurious features are listed as a percentage of models' performance that \textit{depends} on applicability of prediction shortcuts.
Consistently with other figures, the models are ordered based on their absolute average dependency on shortcuts (last column).

\end{document}